\documentclass[12pt]{spieman}  
\usepackage{amsmath,amsfonts,amssymb}
\usepackage{graphicx}
\usepackage{setspace}
\usepackage{tocloft}
\usepackage{textcomp}

\title{Neural Network Pruning by Gradient Descent}

\author[a,b]{Zhang Zhang}
\author[a,b]{Ruyi Tao}
\author[a,b,*]{Jiang Zhang}
\affil[a]{School of Systems Science, Beijing Normal University, Beijing, China}
\affil[b]{Swarma Research, Beijing, China}

\cftpagenumbersoff{figure}
\cftpagenumbersoff{table} 
\begin{document} 
\maketitle

\begin{abstract}
The rapid increase in the parameters of deep learning models has led to significant costs, challenging computational efficiency and model interpretability. In this paper, we introduce a novel and straightforward neural network pruning framework that incorporates the Gumbel-Softmax technique. This framework enables the simultaneous optimization of a network's weights and topology in an end-to-end process using stochastic gradient descent. Empirical results demonstrate its exceptional compression capability, maintaining high accuracy on the MNIST dataset with only 0.15\% of the original network parameters. Moreover, our framework enhances neural network interpretability, not only by allowing easy extraction of feature importance directly from the pruned network but also by enabling visualization of feature symmetry and the pathways of information propagation from features to outcomes. Although the pruning strategy is learned through deep learning, it is surprisingly intuitive and understandable, focusing on selecting key representative features and exploiting data patterns to achieve extreme sparse pruning. We believe our method opens a promising new avenue for deep learning pruning and the creation of interpretable machine learning systems.

\end{abstract}

\keywords{Neural Network Pruning, Gumbel Softmax, Machine Learning Interpretability}

{\noindent \footnotesize\textbf{*}Jiang Zhang,  \linkable{zhangjiang@bnu.edu.cn} }

\begin{spacing}{2}   

\section{Introduction}
\label{sect:intro}  

In the era of large models, deep learning has achieved significant breakthroughs across various fields, including natural language processing\cite{zhao2023survey,min2023recent}, computer vision\cite{kirillov2023segment}, bioinformatics\cite{stahlschmidt2022multimodal}, climate\cite{lam2022graphcast}, medicine\cite{thirunavukarasu2023large}, etc. However, as these models become increasingly complex, they also become over-parameterized. This escalation in parameters leads to not only a sharp rise in both training and inference costs but also a reduction in model interpretability.

As a result, researchers are exploring sparse neural networks to address the issue of over-parameterization in models. Common strategies to achieve network sparsity include network pruning\cite{vadera2022methods}, knowledge distillation\cite{hu2022teacher}, and designing based on prior knowledge\cite{bhoi2021bio}. Among these, network pruning has garnered significant attention due to its good performance, ability to generate flexible sub-network structures and its independence from domain-specific prior knowledge. The core challenge of neural network pruning lies in finding a sub-network that is as small as possible while retaining maximum effectiveness within the original neural network. This task essentially represents an NP-hard combinatorial optimization problem\cite{papadimitriou1998combinatorial}. Consequently, one fundamental approach to tackling this challenge is employing heuristic search methods. One of the earliest papers to explore neural network pruning proposed a method where weights below a certain threshold were pruned during the training process, followed by retraining the remaining network\cite{han2015learning}. This approach laid the foundation for what is now known as \textbf{Iterative Magnitude Pruning (IMP)}. The IMP paradigm involves scoring weights, pruning the lower-scoring ones, and then retraining the network, repeating this cycle. Subsequent research has developed more sophisticated strategies for scoring weights under this paradigm, enhancing model performance. For instance, the concept of connection sensitivity was introduced for evaluating weights\cite{lee2018snip}.  Magnitude of weights with their gradients were combined for scoring the weights\cite{lubana2020gradient,evci2022gradient}. Inspired by bioinformatics, the SynFlow method was invented to score weights by defining synaptic strengths\cite{tanaka2020pruning}. Simultaneously, another class of methods known as \textbf{Dynamic Sparse Training (DST)} has been evolving in parallel. These methods allow the network's topology to evolve during training by adding and pruning connections. The technique became well-known through Sparse Evolutionary Training (SET) method\cite{mocanu2018scalable}, who achieved superior performance compared to static sparse networks. Related works have made numerous improvements in the strategies for growing and pruning connections. For instance, DeepR method introduced strict pruning rules that pruns a weight only if it passes 0 during training\cite{bellec2017deep}, while redistribution strategies for growing new connections were also implemented\cite{mostafa2019parameter,dettmers1907sparse}. RigL modified the network topology using parameter magnitudes and infrequent gradient calculations\cite{evci2020rigging}. However, methods from IMP and DST mainly use stochastic-based or greedy-based growth strategies. The former usually results in lower accuracy, while the latter leads to a suboptimal sparse pattern by introducing inductive biases and greedily searching for structures with local optima, resulting in limited weight coverage. Beyond heuristic search, another promising direction involves treating the network topology as an optimizable variable, known as \textbf{Learnable Sparsity}. Unlike continuously varying weights, the topology consists of discrete variables that cannot be optimized directly through stochastic gradient descent. To address this, various techniques have been proposed to solve this problem indirectly, such as reparameterization trick\cite{louizos2017learning}, continuous relaxation methods\cite{savarese2020winning}, learnable threshold techniques\cite{kusupati2020soft,liu2020dynamic,lee2019differentiable} and estimated gradient for connections\cite{xiao2019autoprune}. These methods enable the optimization of the network topology, but they do not allow for the direct gradient optimization of each weight with under controlled variance. This limitation may often results in suboptimal performance.

To leverage the powerful stochastic gradient descent technique for optimizing neural network structures, a novel approach known as gumbel softmax\cite{jang2016categorical} has captured our attention. Gumbel softmax employs a reparameterization trick, enabling direct gradient-based optimization of categorical distribution parameters. It also offers the flexibility to adjust the variance during the sampling process by controlling the temperature parameter. This method has been widely applied in optimization problems on network science\cite{li2021gumbel} including network reconstruction\cite{zhang2019general,zhang2021automated} and network completion\cite{chen2022inferring}.

To address these challenges in neural network pruning, we introduce a simple yet flexible method for neural network pruning based on gumbel softmax, enabling the direct optimization of the network's topology alongside its weights through stochastic gradient descent. Our contributions are as follows:
\begin{itemize}
    \item We propose a straightforward model that allows the existence probability of each connection in the neural network's topology to be directly optimized by stochastic gradient descent.
    \item Our model demonstrates remarkable compression efficiency. On the MNIST dataset, we have managed to compress the neural network's weights to just 1.5\textperthousand \  of their original size, successfully accomplishing classification tasks with only 404 weights.
    \item The extreme compression of the neural network structure offers interpretability in two aspects: Regarding the learned pruning strategies, we found that it automatically learns smart tricks, including selecting representative features while discarding others, and discovering and utilizing patterns in the data. In terms of understanding the problem itself, our model not only elucidates the importance of features but also allows us to visually observe emergent information pathways in the structure. This reveals which input variables are responsible for which output variables.
\end{itemize}

\section{Related Works}

\subsection{Learable Sparsity}
Optimizing the topology of neural networks directly using gradient descent has captured researchers' attention as a potential solution to the limitations of heuristic algorithms, which often suffer from insufficient heuristic information and extensive computational requirements. Network topology can not be optimized directly by stochastic gradient descent due to its discrete nature.  Researchers have devised various indirect methods to tackle this issue. For instance, one of the earliest well-known works introduced the L0 regularization method\cite{louizos2017learning}, optimizing the existence probability of each variable through the reparameterization trick. However, this method could not control the variance in sampling, potentially leading to significant errors. Another approach employed continuous relaxation to optimize the probability of weights' existence\cite{savarese2020winning}, but its deterministic nature restricted the exploratory range of the network topology. An alternative way is to use  learnable thresholds for joint optimization of structure and parameters\cite{kusupati2020soft,liu2020dynamic,lee2019differentiable}, but threshold-based pruning still encounters the issue of insufficient heuristic information. Optimizing the probability of connection existence through estimated gradients is another viable approach\cite{xiao2019autoprune}, but the accuracy of these estimations is not always guaranteed under various conditions. Our work adopts a similar approach to L0 regularization but incorporates the gumbel-softmax technique\cite{jang2016categorical}. By adjusting the temperature parameter, we can control the variance introduced during the sampling process, achieving performance surpassing other methods.

\subsection{Gumbel Softmax for Optimization Problems}
The gumbel softmax trick is a method for optimizing the parameters of categorical distributions (such as the Bernoulli distribution for binary classification problems) using gradient descent\cite{jang2016categorical}. The essence of this method is a reparameterization technique, which involves introducing random variables to simulate the sampling process. This makes the relationship between the distribution's parameters and the sampling outcome differentiable, allowing for the optimization of the distribution's parameters using stochastic gradient descent. Originally invented for simulating Gaussian distribution sampling in Variational Autoencoders (VAEs)\cite{kingma2013auto}, it has been extended to simulate Bernoulli distribution sampling. This technique has found widespread application in optimizing discrete variables escepially in network science. For example, the GGN method employs gumbel softmax to optimize network structures to fit time-series data on nodes\cite{zhang2019general,zhang2021automated}. It has also been applied to network completion problems where only limited time-series data and partial structure are observable\cite{chen2022inferring}. Moreover, other classic optimization problems in network science, such as modularity maximization problem, maximal independent set (MIS) and minimum vertex cover (MVC) problems can also be effectively solved with gumbel softmax, offering solutions with reduced computational costs\cite{li2021gumbel}.

\section{Neural Network Pruning with Gradient Descent}

\subsection{Preliminary}
Consider a neural network of interest with its parameters divided into two distinct parts: the weights $\theta^w \in \Theta^w \subset \mathbb{R}^d$, and the gating parameters $\theta^g \in \Theta^g \subset (0,1)^d$, where $\Theta^w$ and $\Theta^g$ are the parameter space and $d$ is the number of the model parameters, $\theta^g_i$ denotes the $i$-th entry of $\theta^g$ and it means the probability of $i$-th weight to be retained. $g \in \{0,1\}^d$ denotes the gating variables that sampled from $\theta^g$, where for each $i$, $g_i$ is independently sampled from a gumbel softmax operation with parameter $\theta^g_i$, so we have $g_i \sim gumbel-softmax(\theta^g_i)$ to be either 0 or 1. The standard Hadamard(wlement-wise) product of $\theta^w$ and $g$ is denoted as $\theta^w \odot g$.

\subsection{Model Architecture}
\begin{figure}
\begin{center}
\begin{tabular}{c}
\includegraphics[height=9cm]{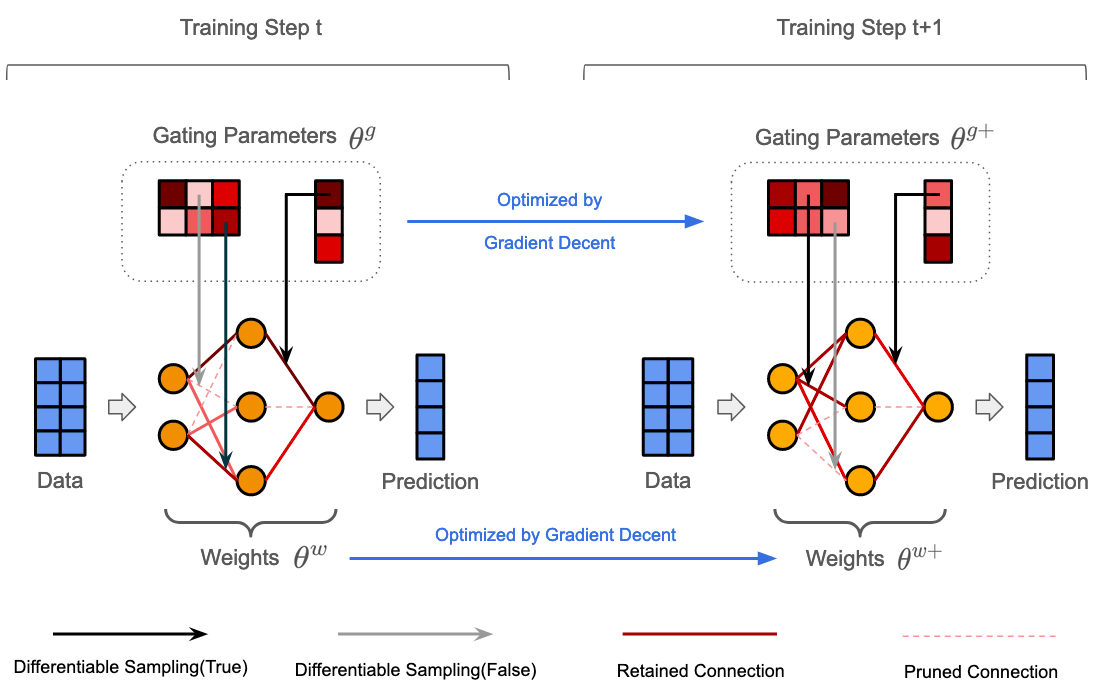}
\end{tabular}
\end{center}
\caption 
{ \label{fig:architecture}
\textbf{A Schematic representation of our framework:}Illustration of the neural pruning process across two consecutive training steps, t and t+1. The network optimizes both gating parameters ($\theta^g$) and weights ($\theta^w$) using gradient descent. The process transitions from differentiable sampling of connections (true or false) to the pruning of selected connections, thereby streamlining the network structure for given machine learning tasks while retaining gradient pathways. } 
\end{figure}

The operation of our model is straightforward and intuitive: before each time of forward propagation, discrete gating variables $g$ are sampled using the gumbel softmax method from a distribution parameterized by $\theta^g$. These gating variables $g$ are then combined with the weight parameters $\theta^w$ through an element-wise production to produce the final weights $\theta^w \odot g$ used in the forward propagation. This approach enables us to directly optimize both the weight parameters $\theta^w$ and the gating parameters $\theta^g$ using stochastic gradient descent. The overall architecture of this process is illustrated in Fig \ref{fig:architecture}.

An important point to note is that the differentiable sampling process in Fig \ref{fig:architecture} is achieved through gumbel softmax. The sampling process from a Bernoulli distribution is typically non-differentiable, which would prevent the passage of gradient information. In our approach, we use the gumbel softmax operation to replace the conventional sampling process. This substitution not only renders the sampling process differentiable but also allows for control over the variance in the sampling process by adjusting the temperature parameter. Specifically, for the parameter $\theta^g_i$, we employ the gumbel sampling through Eq \ref{eq:gumbel_softmax} to obtain the soft gating variable $g_i$.

\begin{equation}
\label{eq:gumbel_softmax}
g_i = \frac{exp((log(\theta^g_i)+\xi_i)/\tau)}{exp((log(\theta^g_i)+\xi_i)/\tau)+exp((log(1-\theta^g_i)+\xi_i^\prime)/\tau)}  ,
\end{equation}

where $\xi_i$ and $\xi_i^\prime$ are i.i.d. random numbers following the gumbel distribution\cite{nadarajah2004beta}. This calculation uses a continuous function with random noise to simulate a discontinuous sampling process. And the temperature parameter $\tau$ adjusts the
sharpness of the output distribution thus adjust the variance. When $\tau \rightarrow 0$, $g_i$ will be 1 with probability $\theta^g_i$ and 0 with probability $1- \theta^g_i$. We can see here that although we can get the hard sample 0 or 1 when $\tau \rightarrow 0$, in practice to maintain the gradient flow $\tau$ can not be really small. Therefore, the trick here is to get the final hard sample result from Eq \ref{eq:hard}

\begin{equation}
\label{eq:hard}
g_i = g_i + \hat{g_i} - \hat{g_i}^\prime ,
\end{equation}

where $\hat{g_i}$ is the soft sample result with a controllable temperature $\tau$ and $\hat{g_i}^\prime$ is the $\hat{g_i}$ who has been detached from its gradient, thus the gradient can be retained with the hard sample results.

\subsection{Loss Function}

Our model's overarching goal is to minimize the number of parameters involved in computations while maintaining the model's capability. Since the active parameters in the model are controlled by $\theta^g$, the loss function of the model is dual-component structured, its combined with the prediction loss and the sparsity loss. The formulation of the loss function is presented in Eq \ref{eq:loss},

\begin{equation}
\label{eq:loss}
L = Prediction\ Loss(\theta^w,\theta^g) + \alpha*|\frac{1}{N}\sum_i^Ngumbel\_softmax(\theta^g_i)-D_{target}|
\end{equation}

where Prediction Loss can vary in different tasks. $\alpha$ means the weight of the sparsity loss and it's an adjustable hyper-parameter. $D_{target}$ is the target density of the model weights. The design of the sparsity loss is based on the idea of sampling the overall density of the network from $\theta^g$, rather than the density of a specific layer. The objective is to ensure that this overall density closely aligns with a predetermined target value. Experimental evidence has shown that such a design is both efficient and flexible. We observe that the neural network automatically adjusts the connection density between different hidden layers to meet computational demands while adhering to the overall density requirement.

\section{Experiments}
In this section, we will demonstrate the training effectiveness and the analysis of our model through a series of experiments. The first subsection will showcase the model's remarkable compression efficiency. In the second and third subsections, we will observe how the compressed model aids in understanding the problem itself. This includes the importance of data features and the emergent information pathways – specifically, how certain input features contribute to certain outputs through well-defined information propagation paths. In the fourth subsection, we will decode the pruning strategy learned by the model. We aim to comprehend how the model solves problems with such a small number of weights. Our findings suggest that the model achieves this by selectively focusing on a few representative features and maximally exploiting patterns present in the data, leading to an optimized reduction in the number of weights.

\subsection{Model Performance}

In this subsection, we will present two experiments to vividly illustrate the capabilities of our model. We aim to show that our approach not only significantly outperforms a random baseline in terms of compression efficiency and retained capability, but it also surpasses competing models.

\begin{figure}
\begin{center}
\begin{tabular}{c}
\includegraphics[scale=0.33]{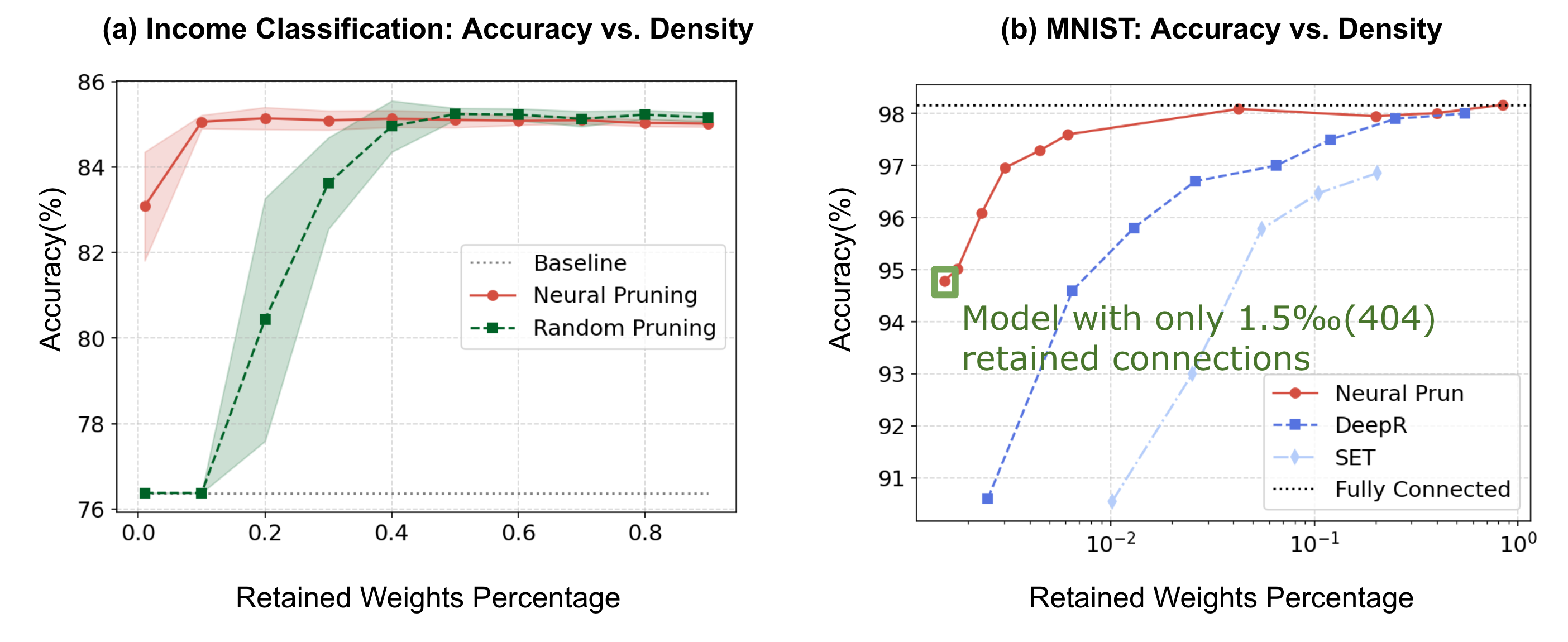}
\end{tabular}
\end{center}
\caption 
{ \label{fig:compare}
\textbf{Comparative Analysis of Pruned Neural Networks:} Performance comparison of the neural pruning framework on two benchmark tasks: (a) Income Classification and (b) MNIST digit recognition. The x-axis represents the percentage of retained weights after pruning, and the y-axis shows the model accuracy. Suplot (a) compares our neural pruning method against baseline and random pruning, while subplot (b) illustrates that our pruned model, with only 0.15\% (404) of the weights retained, not only maintains high accuracy but does so with fewer connections than competing approaches such as DeepR, SET, and fully connected networks.} 
\end{figure}

As depicted in Fig \ref{fig:compare}, we selected widely used datasets in the machine learning field: the Income classification dataset and the MNIST dataset. For the income classification problem, features such as education level are used to predict whether an individual's annual income exceeds 50K. This dataset was chosen for its straightforward feature meanings, aiding in the intuitive analysis of feature importance. As a baseline, we utilized a method that randomly deletes weights while maintaining a fixed network structure. As shown in the left subplot of Fig \ref{fig:compare}, our method significantly outperforms the random deletion approach, maintaining high classification accuracy even with only 1\% of the weights remaining.

The right subplot of Fig \ref{fig:compare} demonstrates the classification and compression performance of different models on the MNIST dataset. Here, we considered a fully connected network with two hidden layers, each containing 300 and 100 neurons, respectively, as a benchmark\cite{lecun1998gradient}. We then pruned this neural network with different method. Classic pruning methods such as DeepR\cite{bellec2017deep} and SET\cite{mocanu2018scalable} were selected as comparative models. Our findings reveal that our method not only achieves higher accuracy than competitors at the same density but also exhibits impressive compression capabilities: the most extensively pruned model retains only 1.5\textperthousand \ of the original model's weights (merely 404 weights) yet still maintains an accuracy rate exceeding 94\%, as shown in the right subplot of Fig \ref{fig:compare}. We were surprised by its compression rate because from common sense, 404 weights seems to be a level of model compression efficiency that is very challenging to surpass in terms of magnitude.

\subsection{Feature Importance from the Pruned Neural Network}

Feature importance is a crucial issue in the field of machine learning interpretability, and common methods for measuring it often require additional computations, such as counterfactual experiments\cite{sundararajan2017axiomatic,lundberg2017unified}. Our intuitive hypothesis is that if a model has been sufficiently pruned, meaning each retained weight is important, then we can directly extract feature importance from the pruned neural network structure, eliminating the need for additional experiments. For example, imagine a feature that contributes nothing to the outcome; in a sufficiently pruned neural network, there would be no weights connecting the neurons of the input layer corresponding to this feature. Based on this rationale, we propose a straightforward method for extracting feature importance from the pruned neural network, as delineated in Eq \ref{eq:fea_im}, 

\begin{equation}
\label{eq:fea_im}
F_{i,j} = \frac{|W_{i,j}|}{\sum_k|W_{k,j}|},
\end{equation}

where $F_{i,j}$ denotes the importance for $i$-th neuron in a previous layer to the $j$-th neuron in the following layer. $W_{i,j}$ denotes the weight, $W_{i,j}$ is 0 if the weight has been pruned. Utilizing Eq \ref{eq:fea_im}, we can determine the importance of any neuron in a preceding layer to any neuron in a subsequent layer. By recursively applying this method, we are able to ascertain the importance of input layer neurons, which correspond to features, in relation to output layer neurons, corresponding to predicted values. Figure 2 illustrates the results of our feature importance analysis for both the Income Classification dataset and the MNIST dataset.

\begin{figure}
\begin{center}
\begin{tabular}{c}
\includegraphics[height=7cm]{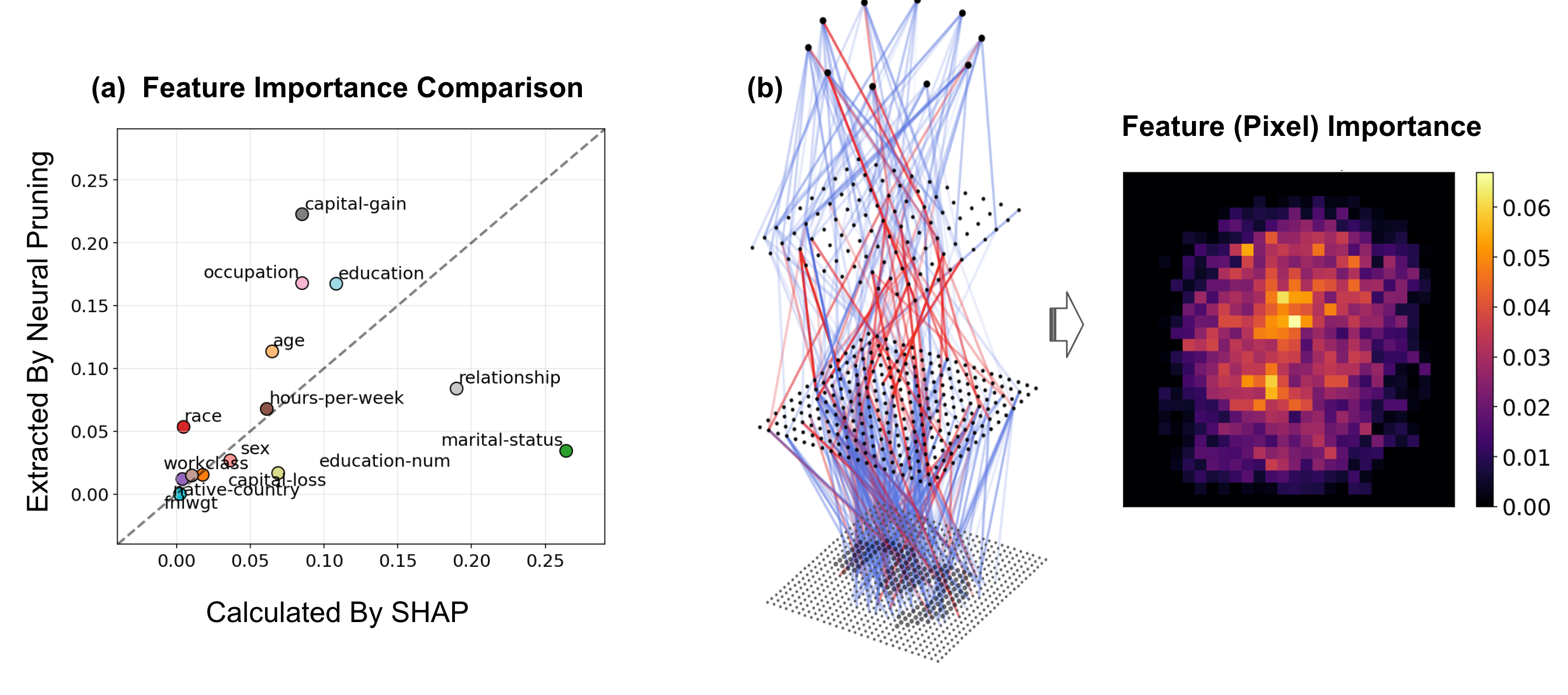}
\end{tabular}
\end{center}
\caption 
{ \label{fig:importance}
\textbf{Feature Importance Analysis in Pruned Networks:} (a) Comparison of feature importance as determined by our neural pruning method (y-axis) and the well-known SHAP method (x-axis), demonstrating a high degree of agreement between the two approaches. (b) Visualization of the pruned neural network structure alongside a heatmap showing the importance of features (pixel positions). The analysis reveals a concentration of significant pixels in the central region of the image, aligning with the intuitive understanding of feature importance in the dataset. } 
\end{figure} 

In the left subplot of Fig \ref{fig:importance}, we visually compare the feature importance measurements provided by our method with those given by the widely acknowledged SHAP method\cite{lundberg2017unified}. It is evident that for most features, the measurements from both methods align closely (points are near the diagonal line), serving as a cross-validation for the accuracy of our feature importance extraction method. It is important to note that while the SHAP method requires counterfactual experiments, our method directly extracts feature importance from the pruned neural network structure, thus offering a significantly faster computational speed than SHAP.

The right subplot of Fig \ref{fig:importance} displays our analysis of feature (pixel) importance for the MNIST dataset. We observe that the important features predominantly cluster in the central area of the images, which aligns with common knowledge as the peripheral areas in MNIST images are typically blank and carry less informative value. 

\subsection{Emergent Information Pathway}

\begin{figure}
\begin{center}
\begin{tabular}{c}
\includegraphics[height=6cm]{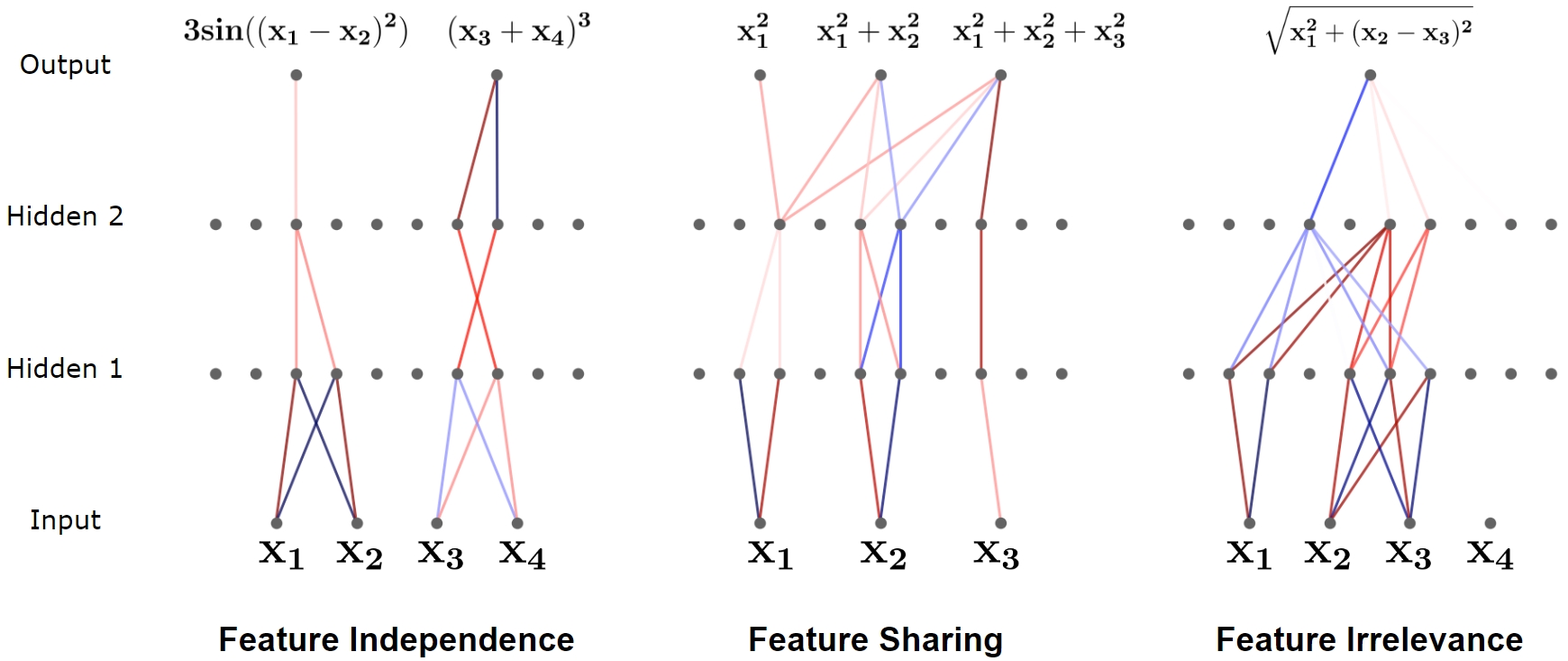}
\end{tabular}
\end{center}
\caption 
{ \label{fig:pathway}
\textbf{
Emergent Information Pathways in Pruned Neural Networks:} This figure displays the topological structures of a thoroughly pruned neural network, where the color of the connections represents the value of the weights. It illustrates the network's ability to directly show the relationships between features and labels across three scenarios: feature independence, feature sharing, and feature irrelevance. Additionally, the symmetry in features is observable, with similar functionalities manifesting equivalent topological structures. } 
\end{figure}

While the understanding of feature importance is undoubtedly crucial for enhancing the interpretability of neural networks, perhaps a more intriguing and valuable question is 'how' these important features affect the labels. In this subsection, we attempt to answer this question by observing emergent information pathways in the sufficiently pruned neural network. In this experiment, we set up several groups of input features, each exhibiting one of three typical relationships with the label data: two groups of input features independently influencing two different labels, various features shared by different labels, and some features being completely unrelated to any label, as illustrated in Fig \ref{fig:pathway}.

In Fig \ref{fig:pathway}, we can observe that our pruning method reveals emergent information pathways, clearly demonstrating how features affect label data under various scenarios: whether the features are independent, shared, or unrelated. In addition to this, feature symmetry can also be easily revealed from the structure of the pruned neural network in Fig \ref{fig:pathway}. We can see that if two features play the same role in the problem, their corresponding local neural network topology is symmetric. For example, $X_1$ and $X_2$, $X_3$ and $X_4$ in Feature Independence task are symmetric respectively. In Feature Irrelevance task $X_2$ and $X_3$ are symmetric. This experiment is inspired by neural network designs driven by prior knowledge of brain networks\cite{liu2023seeing}, where a similar approach is used to create sparse neural networks. We are pleased to note that our method yields pruning results akin to those seen in networks designed with biologically-inspired methods.

\subsection{Understanding the Pruning Strategy}

\begin{figure}
\begin{center}
\begin{tabular}{c}
\includegraphics[height=8.5cm]{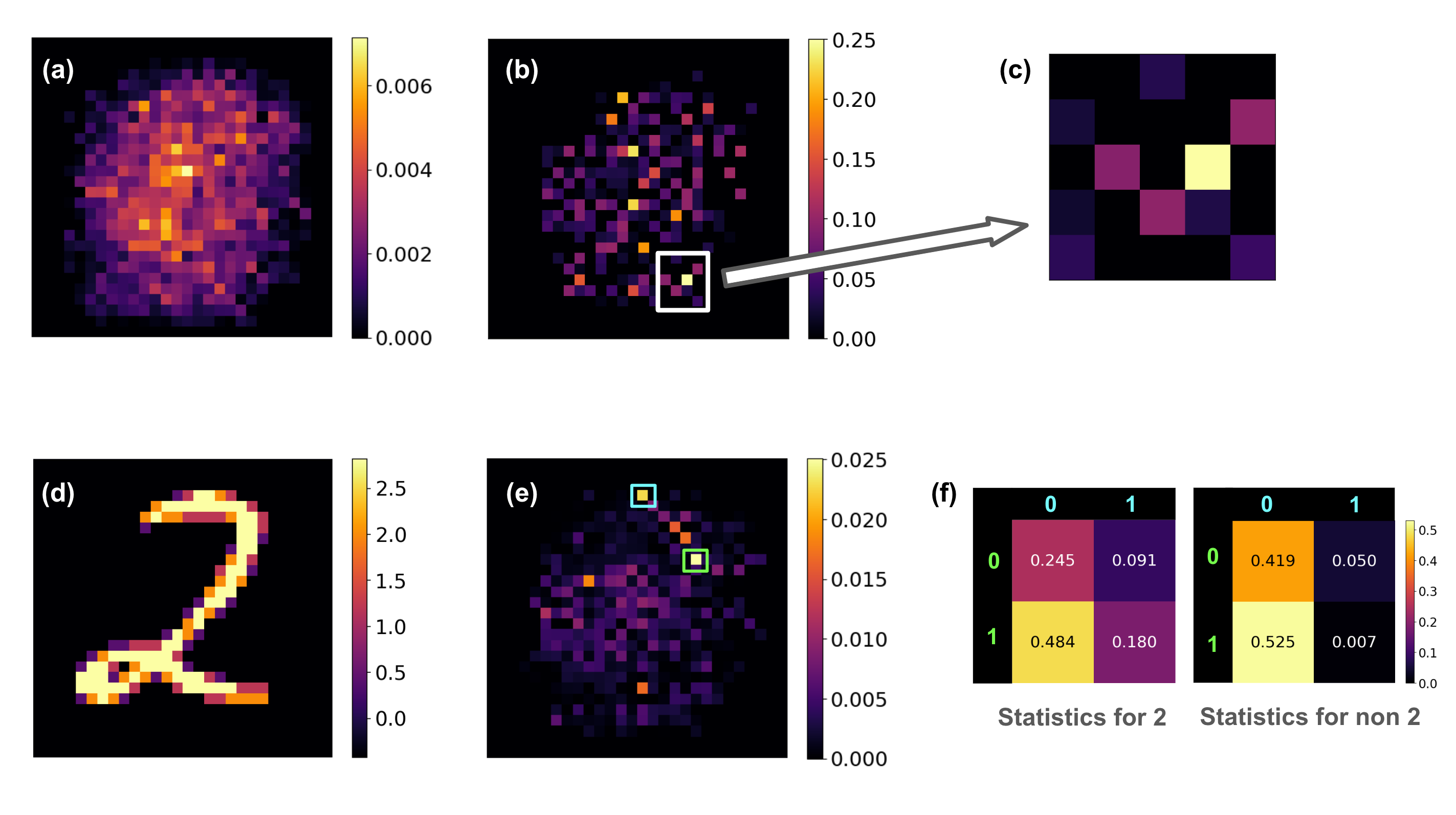}
\end{tabular}
\end{center}
\caption 
{ \label{fig:understand}
\textbf{Learned Pruning Strategy on MNIST Dataset:}  Visualization of feature importance help us reveals the learned pruning strategy,  (a) Overall feature importance with 4\% of connections retained; (b) Overall feature importance with only 0.15\% of connections retained; (c) A zoomed-in view from (b) showing the network's strategy of selecting local, significant feature representatives while discarding others as connections are increasingly pruned. (e) The input feature importance for the output neuron representing the digit '2', highlighting two pixels with exceptionally high importance. (f) The activation distribution of these two pixels across all test set images, showing a distinct difference in the distribution between the images of the digit '2' and others, exemplifying the network's ability to model this distinction using minimal parameters, thereby utilizing effective patterns discovered during the pruning process. } 
\end{figure}

Following the experiments outlined previously, we have observed that our pruning method is highly effective, with the pruning strategy being autonomously learned by the machine. This leads to an intriguing question: what pruning strategy allows the model to maintain its capability while utilizing a minimal number of parameters? To seek answers, we delved into a further analysis of the feature importance results, as depicted in Fig \ref{fig:understand}.

In the upper row of subplots in Fig \ref{fig:understand}, we display the model's feature importance when retaining 4\% and 0.15\textperthousand \   of its original parameters. Notably, in the extremely sparse model, some features (pixels) in the middle region of the image have become significantly less important. In fact, around each important pixel, there are hardly any other important pixels. An analysis of the network structure reveals that most weights corresponding to the neurons of these less important pixels in the input layer have been pruned away. This sheds light on a practical pruning strategy employed by the model under extreme sparsity: \textbf{the selection of pixel representatives}. In the MNIST dataset, adjacent pixels often activate together. Therefore, to maintain an extremely low connection density, the neural network selects to connect only with representative pixels in each area, while neighboring pixels are discarded. 

In the lower row of subplots in Fig \ref{fig:understand}, we illustrate the input feature importance for a specific output label (the example shown is for the digit '2'). We observed that the importance of pixels in two particular positions significantly surpassed that of other areas. To further understand this, we analyzed the distribution of these two pixel positions in data labeled as '2' and those not labeled as '2'. For simplicity, we binarized the data, setting all activated pixels to 1 and non-activated pixels to 0. A clear distribution difference is evident, as shown in Fig \ref{fig:understand}f. This indicates that the activation state of just these two pixels can largely determine the classification of the data. For instance, if both pixels are activated, the model can classify the image as '2' with 74\% accuracy. Theoretically, this determination requires only 2 parameters. This reveals another practical pruning strategy employed by the model under extreme sparsity: \textbf{maximizing the use of existing patterns} in the data. By automatically discovering and utilizing these patterns, our model achieves impressive performance with a minimal number of weights.

\section{Conclusion and Discussion}

In conclusion, this paper presents a novel neural network pruning method that significantly enhances the model's efficiency while maintaining its performance. Our approach, leveraging the Gumbel Softmax principle for learnable sparsity, directly optimizes both the weights and the structural topology of the network using stochastic gradient descent. Various experiments have demonstrated that our method achieves remarkable compression efficiency and retains high accuracy.

Our analysis of feature importance revealed that the pruned network can directly and intuitively indicate which features are important. This contributes greatly to the interpretability of neural networks. Furthermore, by exploring emergent information pathways, we were able to understand how specific features influence the output, adding depth to our knowledge of neural network decision-making processes.

Despite the common perception of deep learning as a 'black box', our findings reveal that the pruning strategy autonomously learned by the model is intuitive and understandable. The pruning strategy focuses on selecting representative features and maximizing the use of existing data patterns. This approach demonstrates the model's ability to make strategic decisions about which connections to retain.

\textbf{Broader Impacts}\ We believe that the principles and techniques applied in this study have broader implications. We suggest a protential pathway towards creating more efficient and interpretable AI systems. Future research could explore the application of these methods in more complex and diverse datasets, as well as their potential in other areas of machine learning and AI. This work contributes to the ongoing efforts in enhancing neural network efficiency and interpretability. By exploring new methods in pruning and feature importance analysis, it hopes to offer valuable insights for further understanding and refining AI systems, potentially aiding their application in practical scenarios.

\subsection* {Code and Data Availability}
The code required to reproduce the experiment can be found at \\ https://github.com/3riccc/neural\_pruning

\subsection* {Acknowledgments}
We are grateful for supporting from Save 2050 Project which is sponsored by Swarma Club and X-Order.\\
We extend our gratitude for the insightful discussions with Yu Liu from the International Academic Center of Complex Systems at Beijing Normal University, which greatly enriched this research.


\bibliography{article}   
\bibliographystyle{spiejour}   





\end{spacing}
\end{document}